%% file: collas2026_conference.tex
\documentclass{article} 
\usepackage{collas2026_conference,times}
\usepackage{easyReview}

\usepackage{multirow}
\usepackage{booktabs} 

\usepackage[noend]{algorithm2e}

\input{math_commands.tex}

\usepackage{hyperref}
\hypersetup{
    colorlinks=true,
    linkcolor=red,
    filecolor=magenta,
    urlcolor=blue,
    citecolor=purple,
    pdftitle={CLIMB: Centroid-Based Hierarchical Memory for Online Continual Self-Supervised Learning},
    pdfpagemode=FullScreen,
}

\title{CLIMB: Centroid-Based Hierarchical Memory\\for Online Continual Self-Supervised Learning}

\author{Julien Lefebvre \\
Universite Claude Bernard Lyon 1, CNRS, INSA Lyon, \\
LIRIS, UMR 5205, 69622 Villeurbanne, France \\
\texttt{julien.lefebvre@liris.cnrs.fr} \\
\And
Stefan Duffner \\
INSA Lyon, CNRS, Universite Claude Bernard Lyon 1, \\
LIRIS, UMR 5205, 69621 Villeurbanne, France \\
\texttt{stefan.duffner@liris.cnrs.fr} \\
\AND
Mathieu Lefort \\
Universite Claude Bernard Lyon 1 / Univ. Rennes, Inria, CNRS\\
LIRIS UMR 5205 / IRISA UMR 6074, France \\
\texttt{mathieu.lefort@liris.cnrs.fr}
}

\collasfinalcopy 

\begin{document}

\maketitle

\begin{abstract}

Online Continual Self-Supervised Learning (OCSSL) aims to learn representations from a continuous stream of unlabeled data, without knowledge of task boundaries and under memory constraints. Existing methods rely either on replay buffers that exploit latent space structure, or on regularization alone. We present CLIMB (\textit{Continual Learning with Intelligent Memory Bank}), which combines both simultaneously. Our method introduces a hierarchical centroid-based memory, bounded in total number of stored images, combined with knowledge distillation on replayed examples to limit representation drift. The memory groups similar images into centroids, providing hard-to-discriminate examples for contrastive learning while covering the diversity of observed distributions. Experiments on Split CIFAR-100 and Split ImageNet-100, on standard benchmarks from the state-of-the-art as well as a new protocol with irregular task distributions show that CLIMB outperforms state-of-the-art OCSSL methods. Source code is available at : \url{https://github.com/lefebvju/climb}

\end{abstract}

\section{Introduction}
\label{sec:intro}
 \textit{Self-Supervised Learning} (SSL) has established itself as the dominant paradigm for representation learning from unlabeled data~\citep{chenSimpleFrameworkContrastive2020,grillBootstrapYourOwn2020,chenExploringSimpleSiamese2020}. SSL methods exploit the intrinsic structure of data through augmentations. Contrastive approaches \citep{chenSimpleFrameworkContrastive2020} attract different views of the same image while repelling those of different images. Non-contrastive approaches~\citep{grillBootstrapYourOwn2020,chenExploringSimpleSiamese2020} prevent representational collapse through architectural asymmetry without requiring negative examples. However, most SSL methods assume that all training data is jointly available, which is not realistic when data evolves over time. In such settings, these systems either require full retraining from scratch at each update, or suffer from \textit{catastrophic forgetting}, the tendency of neural networks to overwrite weights associated with past tasks when learning new ones~\citep{kirkpatrickOvercomingCatastrophicForgetting2017}. The domain of \textit{Continual Learning} (CL) addresses this challenge by enabling models to acquire new knowledge incrementally without erasing prior representations~\citep{wangComprehensiveSurveyContinual2024}.
 
A sub-branch of continual learning focuses on unannotated data, making it possible to exploit far larger data volumes by removing the need for labels. The intersection of CL and SSL defines \textit{Continual Self-Supervised Learning} (CSSL), an emerging field explored by several recent works~\citep{finiSelfSupervisedModelsAre2022a,gomez-villaContinuallyLearningSelfSupervised2022,zhangIntegratingPresentUnsupervised2024}. Generally in CSSL, methods assume that multiple epochs can be performed on each task and that task boundaries are known, allowing catastrophic forgetting mitigation mechanisms to be triggered explicitly at each transition. We focus on a subdomain of CSSL, \textit{Online Continual Self-Supervised Learning} (OCSSL)~\citep{yuSCALEOnlineSelfSupervised2023}, where the model performs a single pass over the incoming data stream using mini-batches, with no access to task boundaries. This setting is particularly relevant for applications requiring rapid adaptation and continuous representation evaluation, for example an autonomous agent exploring a novel environment. Fair comparison between methods requires controlling both memory capacity and computational budget simultaneously, a constraint barely addressed in the existing literature.
 
In this paper, we present \textbf{CLIMB}, which addresses the OCSSL setting by combining a novel strictly bounded hierarchical centroid-based memory with a classical EMA-based regularization. Our contributions are:
\begin{enumerate}
    \item \textbf{A hierarchical centroid-based memory} inspired by PCMC~\citep{taylorPATCHBASEDCONTRASTIVELEARNING2024}, compatible with the OCSSL setting under strict memory constraints. Unlike PCMC, CLIMB operates with full-image representations rather than patches, removes both the offline pretraining phase and the sleep phases, and imposes a strict global bound on memory size. Combined with knowledge distillation on replayed examples following CLA~\citep{cignoniCLALatentAlignment2025}, this memory improves performance in most configurations, with a particularly pronounced advantage when data is observed in a more fragmented fashion, i.e., when each task covers a smaller subset of classes.
 
    \item \textbf{A new evaluation protocol with irregular task distributions} for the OCSSL setting, where the number of classes per task varies randomly, following irregular distributions previously explored in supervised continual learning~\citep{koh2022online}, in order to measure the robustness of OCSSL methods against more varied continual learning configurations than the balanced tasks usually considered.
\end{enumerate}

The related work is presented in Section~\ref{sec:related}, followed by a detailed description of our approach (Section~\ref{sec:method}). Experimental protocols and analysis of results are then developed in Sections~\ref{sec:exp} and~\ref{sec:results}, before concluding and discussing perspectives in Section~\ref{sec:conclusion}.


\section{Related Work}
\label{sec:related}

\label{sec:related_CL}

In continual learning, three main families of methods have been developed.

\textbf{Regularization-based methods} constrain network weight updates based on their importance to previous tasks. EWC~\citep{kirkpatrickOvercomingCatastrophicForgetting2017} exploits the Fisher information matrix to estimate the importance of each parameter and penalizes their modification when learning new tasks. SI~\citep{zenkeContinalLearningthrough2017} extends this idea by evaluating parameter importance along the full learning trajectory. Both methods operate in a supervised setting. A second regularization family targets model outputs rather than parameters, using a frozen previous model as a teacher via a distillation loss. LwF~\citep{liLearningForgetting2017} is the founding work of this approach. CaSSLe~\citep{finiSelfSupervisedModelsAre2022a} adapts it to the offline self-supervised setting via a dedicated projection head.

\textbf{Architecture-based methods} dynamically adapt model structure to allocate capacity to new tasks. Progressive Neural Networks~\citep{rusuProgressiveNeuralNetworks2022} dedicate a new backbone to each task while adding lateral connections to reuse past representations. U-TELL~\citep{solomonUTELLUnsupervisedTask2024} extends this paradigm to the unsupervised setting by dedicating a new expert module to each incoming task.

\textbf{Replay-based methods} revisit past data during training. They divide into two main categories: generative approaches, which use a generative model to synthesize past examples without requiring explicit storage of input~\citep{cywinskiGUIDEGuidancebasedIncremental2024,solomonUTELLUnsupervisedTask2024}, and episodic memory approaches, which maintain a buffer containing a subset of past data for replay during training~\citep{purushwalkamChallengesContinuousSelfSupervised2022, taylorPATCHBASEDCONTRASTIVELEARNING2024, cignoniCLALatentAlignment2025}. These approaches have been developed in both supervised~\citep{cywinskiGUIDEGuidancebasedIncremental2024} and self-supervised settings~\citep{purushwalkamChallengesContinuousSelfSupervised2022,solomonUTELLUnsupervisedTask2024,cignoniCLALatentAlignment2025}.

These three families are not mutually exclusive and are frequently combined to leverage their complementarity. In a supervised setting, methods such as iCaRL~\citep{rebuffiICaRLIncrementalClassifier2017} combine replay and distillation to better preserve past knowledge. In a self-supervised setting, replayed examples are also frequently combined with knowledge distillation regularization, where a frozen previous model constrains current representations from drifting excessively~\citep{finiSelfSupervisedModelsAre2022a,gomez-villaContinuallyLearningSelfSupervised2022,cignoniCLALatentAlignment2025}. U-TELL~\citep{solomonUTELLUnsupervisedTask2024} combines architectural modifications with generative replay in an unsupervised setting.

\label{sec:related_ocssl}
Online supervised continual learning has been extensively studied~\citep{buzzega2020dark, koh2022online, caccia2022new}, where the single-pass constraint requires updating representations from each mini-batch only once without storing the full dataset. Online Continual Self-Supervised Learning (OCSSL) extends this challenge to the unlabeled setting, combining the absence of annotations, a single-pass data stream, and unknown task boundaries. Existing methods augment a standard SSL backbone with mechanisms from the families described above: a regularization loss constraining representation drift, an episodic memory for replaying past examples, or a combination of both.

Comparing OCSSL methods equitably is non-trivial, as performance depends on both memory capacity and computational budget. A method with access to more stored examples benefits from more diverse replay, while a method performing more gradient updates exploits that diversity more effectively. Without controlling both simultaneously, comparisons become meaningless: a well-organized memory paired with a larger budget would trivially outperform a simpler method under tighter constraints. CLA~\citep{cignoniCLALatentAlignment2025} identifies this open problem and introduces Cumulative Backward Passes (CBP) as a principled metric, representing the total number of images that undergo backpropagation during training. We will adopt this metric throughout our experiments to ensure fair comparison across all methods.

Replay-based methods maintain  a buffer of past examples used during training. Minred~\citep{purushwalkamChallengesContinuousSelfSupervised2022} keeps a subset of maximally decorrelated examples, discarding the most redundant at each insertion, and uses only memory examples in the SSL loss without additional regularization. Osiris-R~\citep{zhangIntegratingPresentUnsupervised2024} uses two parallel projection heads: one dedicated to plasticity, optimized on current stream examples, and one dedicated to cross-task consolidation, optimized on both stream and replay examples via a contrastive loss. Memory relies on reservoir sampling~\citep{vitterRandomSamplingReservoir1985} for the buffer. PCMC~\citep{taylorPATCHBASEDCONTRASTIVELEARNING2024} takes a
different approach, maintaining a centroid-based memory that approximates the distribution of data representations, storing raw image patches to avoid representation drift due to encoder updates. Clusters of similar patches in representation space are distributed between a short-term memory (STM) and a long-term memory (LTM). The method proceeds in two phases: a wake phase during which the model generates clusters in the STM and fills them with patches. In the middle of each task, a sleep phase where the encoder is retrained in offline SSL mode on patches stored in STM and LTM, then centroids are recalculated with the new encoder to reposition them in latent space. PCMC's hierarchical centroid structure represents a promising approach to organizing replay memory, motivating its adaptation to the OCSSL setting. Although PCMC presents itself as an OCSSL method, several design choices imply substantially different constraints: the LTM grows without bound, and sleep phases involve 200 epochs of offline training on the entire memory, making CBP computation impossible since the memory has no defined size. Furthermore, since memory capacity is unbounded, optimizing the method effectively amounts to maximizing the number of stored images through hyperparameter tuning. PCMC also requires an initial offline pretraining task, constituting prior knowledge absent in other OCSSL methods. These differences make PCMC incompatible with the two constraints we impose for fair comparison, a fixed memory capacity and a measurable CBP budget.

Methods combining replay with regularization constrain representation drift in addition to replaying past examples. SCALE~\citep{yuSCALEOnlineSelfSupervised2023} organizes memory via the Part and Select Algorithm~\citep{salomonPSANewScalable2013} to maximize representational diversity, and penalizes divergence between the current model's similarity matrix and that of the previous training step. CLA~\citep{cignoniCLALatentAlignment2025} also combines replay with a distillation loss, but proposes a more stable reference by aligning current representations with past ones on replayed examples only, preserving plasticity for new data. Two variants are proposed: CLA-E uses an EMA encoder updated after each mini-batch as a stable distillation reference, while CLA-R stores embeddings from the first pass through the encoder directly as alignment targets. CLA achieves state-of-the-art performance among the OCSSL methods evaluated in their paper, with CLA-E and CLA-R each leading on different metrics and configurations.

Faced with these observations, we propose CLIMB, which builds on the idea of a hierarchical centroid-based memory to replace the FIFO buffer of CLA, the current state-of-the-art OCSSL method. Rather than retaining only the most recent examples as CLA does, CLIMB maintains a structured memory that covers the entire stream, raising the question of what to preserve and discard under a fixed capacity constraint, while preserving CLA's distillation mechanism as a complementary component to limit representation drift.

\section{Method}
\label{sec:method}

\subsection{Architecture}
\label{sec:archi}

\begin{figure}[t]
    \centering
    \includegraphics[width=0.5\textwidth]{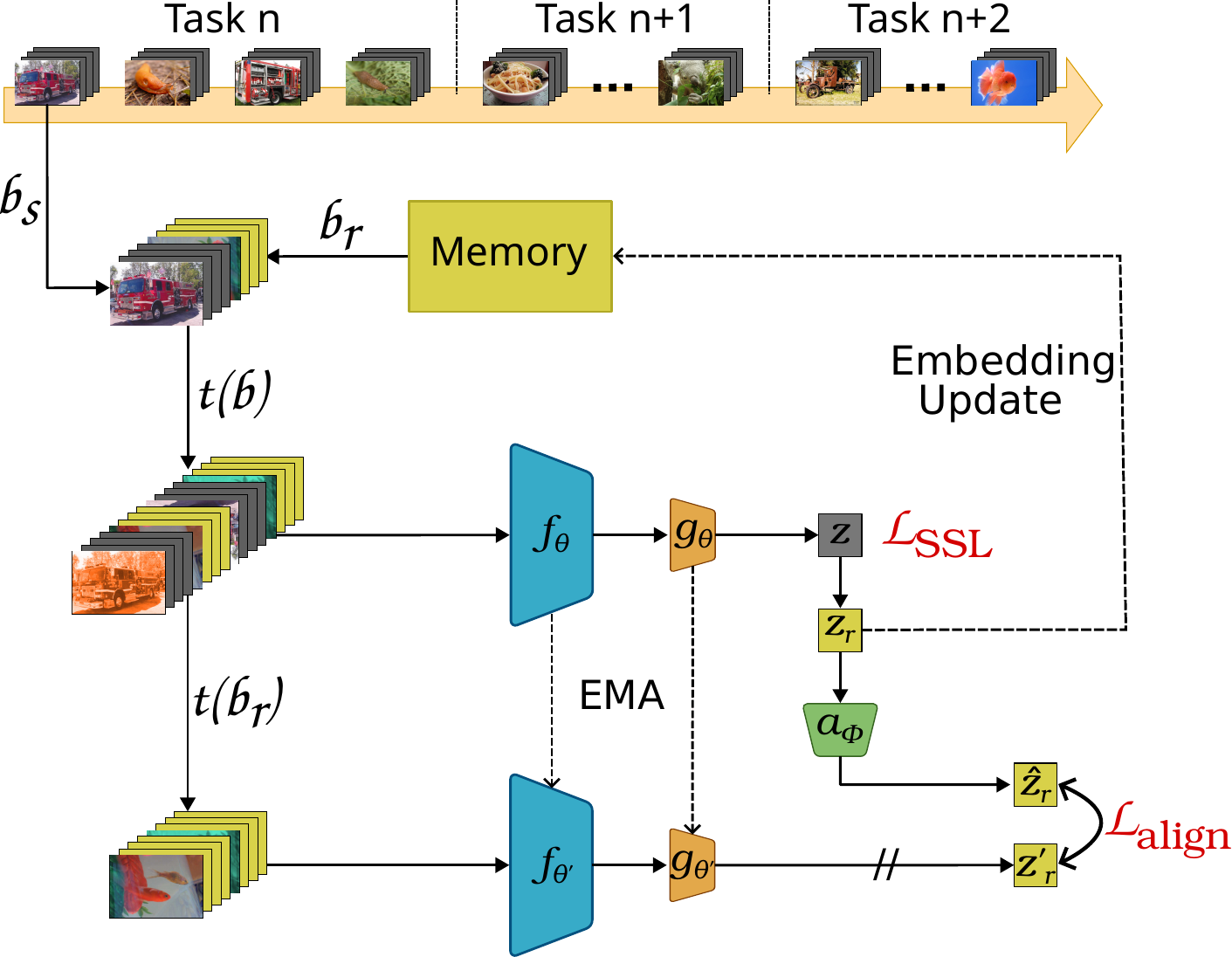}
    \caption{Overview of CLIMB's architecture. At each step, a stream mini-batch $b_s$ is combined with a replay batch $b_r$ sampled from the hierarchical centroid memory to form the final batch $b = b_s \cup b_r$. The online network ($f_\theta$, $g_\theta$) processes $b$ under two augmented views, producing embeddings $z = g_\theta(f_\theta(t(b)))$ for the contrastive loss $\mathcal{L}_{\text{SSL}}$. The alignment loss $\mathcal{L}_{\text{align}}$ is computed as the negative cosine similarity between present representations of the replay subset, passed through projection head $a_\phi$, and past representations $z'_r = g_{\theta'}(f_{\theta'}(t(b_r)))$ produced by the frozen EMA target network ($f_{\theta'}$, $g_{\theta'}$). Embeddings of replay examples $z_r$ are also used to update the centroid positions in memory.}
    \label{fig:learning}
\end{figure}

CLIMB builds on a standard SSL architecture comprising a backbone $f_\theta$ and a projection head $g_\theta$ optimizing a self-supervised pretext task, as illustrated in Figure~\ref{fig:learning}. Two mechanisms to limit forgetting are integrated: a memory module for replaying past examples, and a knowledge distillation loss using a frozen EMA model as reference to constrain representation drift. These mechanisms thus combine two of the main families identified in the previous Section~\ref{sec:related_CL}: memory and regularization.

During training, each stream mini-batch $b_s$ is encoded by the backbone and projection head, and subjected to a novelty detection step to update the hierarchical memory (STM/LTM, Section~\ref{sec:memory}). A replay batch $b_r$ is sampled from memory and concatenated with the current batch to form the final batch $b = b_s \cup b_r$, used for SSL optimization, while knowledge distillation is applied on $b_r$ only (Section~\ref{sec:training}).

\subsection{Memory}
\label{sec:memory}

CLIMB's memory is a centroid-based structure that maintains a compact yet representative summary of the observed latent space under a strictly bounded capacity. Each centroid groups up to $M$ raw images and is represented in latent space by the exponential moving average of their embeddings. The underlying intuition is that close neighbors in latent space are hard to discriminate under the contrastive loss and likely belong to the same mode of the feature distribution, grouping them into a shared centroid therefore yields a set of anchors that jointly cover the diversity of observed data throughout the stream. Distances are computed in the projected space after $g_\theta$, i.e., the space in which $\mathcal{L}_{\text{SSL}}$ is optimized, ensuring coherence between the centroid structure and the space in which the model learns to discriminate embeddings: hard examples identified by clustering are directly relevant to the training loss.Rather than maximizing raw diversity, CLIMB organizes its memory into semantically coherent and well-separated clusters that jointly cover the diversity of the observed stream, a quantitative analysis of this structure relative to alternative buffer strategies is provided in Appendix~\ref{app:diversity}.

\begin{figure}[t]
    \centering
    \includegraphics[width=0.7\textwidth]{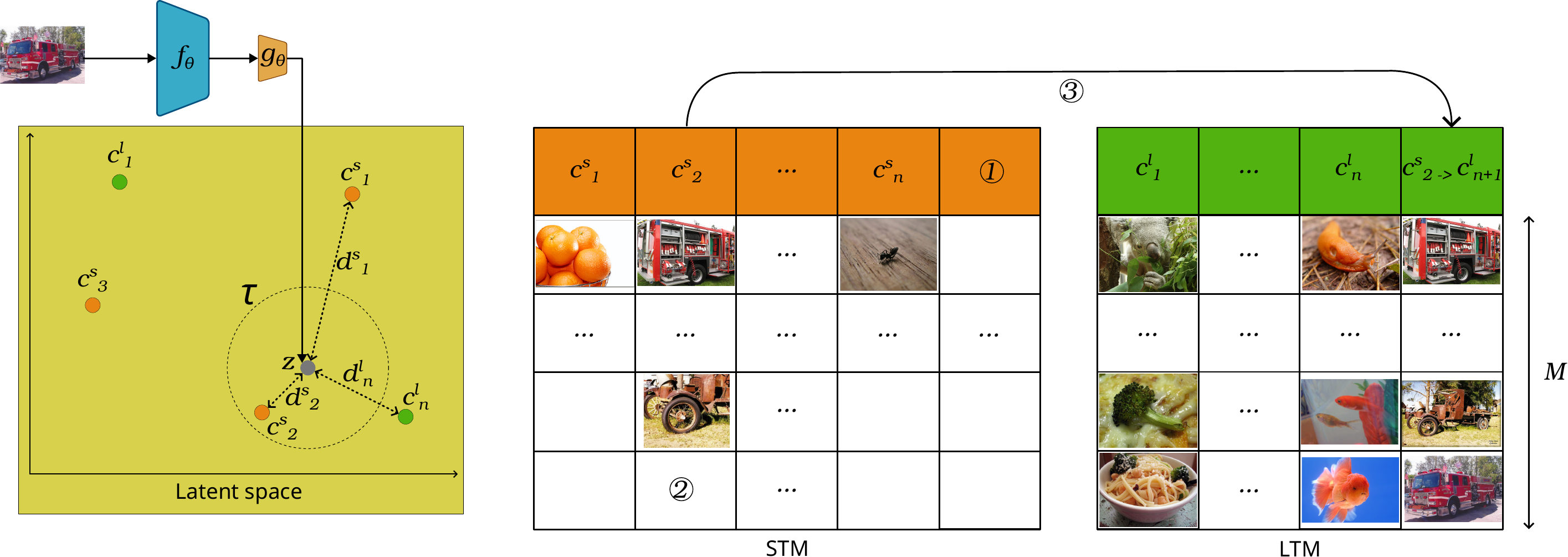}
    \caption{Overview of CLIMB's hierarchical memory. $c_i$ denotes centroid positions in the projected latent space, and $d_i$ denotes the cosine distances between each centroid and the current image embedding. The short-term memory (STM) is shown in orange and the long-term memory (LTM) in green. \textcircled{1}~If $\min(d_i) > \tau$, a new centroid is instantiated from the current image. \textcircled{2}~Otherwise, the image is assigned to the nearest centroid if it belongs to the STM. \textcircled{3}~When a centroid reaches $M$ examples, it is promoted to the LTM and removed from the STM.}
    \label{fig:memory}
\end{figure}

Inspired by the hierarchical designs of STAM~\citep{smithUnsupervisedProgressiveLearning2021} and PCMC~\citep{taylorPATCHBASEDCONTRASTIVELEARNING2024}, memory is split into a short-term memory (STM, up to $L$ centroids) acting as a staging area for newly discovered concepts, and a long-term memory (LTM, up to $K$ centroids) consolidating mature, well-populated centroids (Figure~\ref{fig:memory}). Our main contribution on the memory side, relative to PCMC, is to strictly bound total capacity, both in number of centroids and in number of stored images, while preserving a diverse, up-to-date summary of the stream. The complete update procedure is given in Algorithm~\ref{alg:climb_memory}.

Novelty is detected via an adaptive threshold $\tau$, recomputed at each step as the $p$-th percentile ($p=0.95$) of the last $w=1000$ observed minimum distances, this removes the need to hand-tune a fixed distance cutoff and lets the threshold track the geometry of the representation as it evolves. A sample whose minimum cosine distance to all existing centroids exceeds $\tau$ instantiates a new STM centroid, replacing the least recently updated one if the STM is full. Otherwise the sample is assigned to its nearest centroid. STM assignments accumulate toward the promotion threshold $M$ and update the centroid value via EMA with a fixed coefficient $\alpha_{\text{stm}}$. LTM assignments are accepted with probability $0.5$, in which case they replace a random example in the matched centroid, allowing the LTM to incorporate recent content under bounded storage. When an STM centroid reaches $M$ examples, it is promoted to the LTM. If the LTM is full, the two most similar centroids are merged and $M$ examples are retained by random selection, bounding the number of long-term centroids while preserving latent-space coverage.

Two additional mechanisms enforce overall capacity control. First, a global pruning is triggered when the total number of stored images exceeds $N$: all STM examples are deleted except one anchor per centroid, the image that triggered its creation, retained so that the centroid remains a valid landmark and can accommodate similar incoming samples if they reappear later in the stream. An ablation of this pruning strategy is provided in Appendix~\ref{app:trim}. This mechanism is distinct from LTM merging and targets image count rather than centroid count. Second, centroid values are updated after each gradient step using, for each replayed example, the mean of its two augmented-view embeddings produced by the updated encoder, ensuring that the memory structure remains aligned with the current state of the encoder as representations drift during training. A detailed analysis of these dynamics over a representative training run is provided in Appendix~\ref{app:dynamics}.

\begin{algorithm}[t]
\caption{CLIMB Memory Update (per stream mini-batch)}
\label{alg:climb_memory}
\DontPrintSemicolon
\KwIn{Stream mini-batch: images $x_i$ and their projected embeddings $z_i = g_\theta(f_\theta(x_i))$}
\BlankLine
\textbf{Step 1 --- Nearest centroid.}\;
\ForEach{image $x_i$ in the mini-batch}{
    Compute the cosine distance between $z_i$ and every centroid in STM and LTM.\;
    Record the closest centroid $j^{\star}$ and its distance $d_i^{\star}$.\;
}
\BlankLine
\textbf{Step 2 --- Assignment.}\;
\ForEach{image $x_i$ in the mini-batch}{
\eIf{$d_i^{\star} > \tau$ \textnormal{(novel sample)}}{
    \If{\textnormal{STM is full (}$|\textnormal{STM}| = L$\textnormal{)}}{
        Remove the least recently updated STM centroid and its associated images.\;
    }
    Instantiate a new centroid with value $z_i$ and associated image~$x_i$.\;
}{\eIf{$j^{\star} \in \textnormal{STM}$}{
    Add $x_i$ to the centroid's examples (up to $M$).\;
    Update its value via EMA: $\mu_{j^{\star}} \leftarrow (1-\alpha_{\text{stm}})\,\mu_{j^{\star}} + \alpha_{\text{stm}}\,z_i$.\;
}{
    With probability $0.5$, add $x_i$ to the LTM centroid's examples, replacing a randomly chosen example.\;
}}
}
\BlankLine
\textbf{Step 3 --- Threshold update.}\;
Append the distances $d_i^{\star}$ to a sliding window of size $w$, and recompute $\tau$ as the $p$-th percentile of the window.\;
\BlankLine
\textbf{Step 4 --- STM $\to$ LTM promotion.}\;
\ForEach{STM centroid $j^{\star}$ that received an example in Step~2}{
    \If{$|\mathcal{E}_{j^{\star}}| = M$}{
        Promote $j^{\star}$ to the LTM with its associated images; free its STM slot.\;
        \If{$|\textnormal{LTM}| > K$}{
            Merge the two LTM centroids with highest cosine similarity: pool their examples and retain $M$ by random selection.\;
        }
    }
}
\BlankLine
\textbf{Step 5 --- Memory pruning.}\;
\If{\textnormal{total stored images} $> N$}{
    Delete all STM examples except the first one per centroid, retained as an anchor.\;
}
\BlankLine
\textbf{Step 6 --- Centroid value update.}\;
After the gradient step, update each centroid $j$ that contributed a replay image via EMA with coefficient $\alpha_j = 0.5 / |\mathcal{E}_j|$, using the mean embedding of the two augmented views of that image.\;
\end{algorithm}

\subsection{Replay-Based Knowledge Distillation}
\label{sec:training}

As illustrated in Figure~\ref{fig:learning}, for each stream mini-batch $b_s$ of size $|b_s|$, a replay batch $b_r$ of size $|b_r|$ is sampled in equal parts from the LTM and the STM, forming the final batch $b = b_s \cup b_r$. This batch is transformed according to the augmentations of the chosen SSL method, then passed through the backbone $f_\theta$ followed by projection head $g_\theta$ to compute loss $\mathcal{L}_{\text{SSL}}$. We use SimCLR~\citep{chenSimpleFrameworkContrastive2020} as the base SSL method, whose loss enforces similarity between representations of two augmented views of the same image while pushing apart those of different images in the batch using the same encoder for both views. SimCLR has been extensively studied in the CSSL literature~\citep{finiSelfSupervisedModelsAre2022a, cignoniCLALatentAlignment2025, taylorPATCHBASEDCONTRASTIVELEARNING2024, zhangIntegratingPresentUnsupervised2024}, enabling direct comparison with existing methods.

To limit catastrophic forgetting without task boundaries, an alignment loss is computed on examples from $b_r$, following CLA~\citep{cignoniCLALatentAlignment2025}. The intuition is to constrain representations produced by the current model to remain aligned with those produced by a reference model, preventing excessively rapid drift of the latent space. This constraint is complementary to CLIMB's structured replay: whereas CLA uses a FIFO buffer that retains only the most recent examples, CLIMB's hierarchical centroid memory maintains examples representative of the entire stream, enabling replay to combat forgetting over the long term rather than only in the short term. The reference model is implemented as an EMA encoder $f_{\theta'}$ and its projector $g_{\theta'}$, updated after each mini-batch as $\theta' \leftarrow \tau_{\text{ema}} \theta' + (1 - \tau_{\text{ema}}) \theta$, where $\tau_{\text{ema}} \in [0, 1]$ controls the update speed. A dedicated projection head $a_\phi$ is applied to $z_r$ to align them with representations $z^{\text{ema}}_r = g_{\theta'}(f_{\theta'}(b_r))$ produced by the EMA model. The alignment loss is defined as the negative average cosine similarity between $a_\phi(z_r)$ and $z^{\text{ema}}_r$:

\begin{equation}
\mathcal{L}_{\text{align}} = -\frac{1}{|b_r|} \sum_{i=1}^{|b_r|}
\frac{a_\phi(z_i) \cdot z^{\text{ema}}_i}
     {\|a_\phi(z_i)\| \, \|z^{\text{ema}}_i\|}
\label{eq:loss_forget}
\end{equation}

where $z_i = g_{\theta}(f_\theta(t(x_i)))$ is the representation of replay example $x_i \in b_r$, $a_\phi(z_i)$ is the projection head applied to the representation of $x_i$, and $z^{\text{ema}}_i = g_{\theta'}(f_{\theta'}(t(x_i)))$ its representation from the EMA model. The total loss is then:

\begin{equation}
\mathcal{L} = \mathcal{L}_{\text{SSL}} + \lambda \mathcal{L}_{\text{align}}
\label{eq:loss_total}
\end{equation}

where $\lambda$ is a hyperparameter controlling the weight of the alignment loss relative to the SSL loss.


\section{Experiments}
\label{sec:exp}

\subsection{Datasets and Task Configurations} 

Experiments are conducted on Split CIFAR-100~\citep{krizhevsky2009learning} and Split ImageNet-100~\citep{deng2009imagenet}, the reference benchmarks used in CLA~\citep{cignoniCLALatentAlignment2025} and CaSSLe~\citep{finiSelfSupervisedModelsAre2022a}. All experiments use a \textit{class-incremental} protocol. 

\paragraph{Regular task distribution.} Following~\citet{cignoniCLALatentAlignment2025}, the standard configuration uses 20 tasks of 5 classes drawn randomly per seed, ensuring that all methods face identical class sequences for a given seed. To evaluate method behavior under more challenging conditions, configurations with 50 and 100 tasks are also tested, reducing the number of classes per task while preserving the same total number of classes. This increases task fragmentation, making classes seen in early tasks more susceptible to forgetting. 

\paragraph{Irregular task distribution.} Inspired by~\citet{koh2022online}, 
who introduced irregular task distributions in the supervised continual 
learning setting, to evaluate the robustness of methods against varied task sequences, we conduct experiments with irregular distributions of the number of classes per task, on the 20-task and 50-task configurations. The 100-task configuration is excluded as it would force exactly one class per task, preventing any irregularity. For the 20-task (resp. 50-task) configuration, the number of classes per task is drawn randomly between 1 and 12 (resp. 5), subject to the constraint that the sum of classes across all tasks remains equal to the total number of classes in the dataset. Since task distribution introduces an additional factor of variability beyond model initialization, 20 seeds are used for these experiments, each controlling both the class-to-task assignment, the number of classes per task and model initialization. Full distributions across all seeds are provided in Appendix~\ref{app:distributions}.

\subsection{Models and Evaluation Protocol} 
All experiments use a ResNet-18 encoder~\citep{he2016deep}. The computational budget is measured by CBP (\textit{Cumulative Backward Passes}) as defined by~\citet{cignoniCLALatentAlignment2025}:
\begin{equation}
    \text{CBP} = n_v \times n_{\text{steps}} \times b, \qquad
    n_{\text{steps}} = n_p \times \frac{|\mathcal{D}|}{b_s},
    \label{eq:cbp}
\end{equation}
where $n_v = 2$ is the number of views required by SimCLR, $b = b_s + b_r$ is the total mini-batch size, $n_p$ is the number of sequential passes per incoming mini-batch, and $|\mathcal{D}|$ is the total number of training samples. We follow the high CBP setting of CLA~\citep{cignoniCLALatentAlignment2025}, with $b_s = 10$, $b_r = 128$ and $n_p = 3$, yielding $\text{CBP} = 3.7 \times 10^6$ for Split CIFAR-100 and $\text{CBP} = 9.5 \times 10^6$ for Split ImageNet-100. This budget is maintained equivalent across all compared methods. The total memory capacity is likewise fixed at $N = 2500$ images for all methods. For regular task distribution and ablation experiments, 5 runs with fixed seeds are conducted. For irregular task distribution experiments, 20 seeds are used as described in previous section. 

CLIMB is compared against MinRed, Osiris-R, SCALE, CLA-E, and CLA-R. Results with SimSiam as an alternative base SSL method are reported in Appendix~\ref{app:simsiam}. PCMC~\citep{taylorPATCHBASEDCONTRASTIVELEARNING2024} is included as an indicative reference without its offline pretraining phase, the only modification made to ensure a minimal basis for comparison. Even in this reduced setting, its effective CBP remains substantially higher than the budget used for all other methods due to its offline sleep phases, and its memory grows without bound, reaching between 6000 and 15000 stored patches. Note that these patches are $60 \times 60$ crops rather than full images, making direct memory capacity comparison with other methods difficult.

Performance is evaluated via a linear classifier trained on representations produced by the frozen encoder $f_\theta$ on all classes seen so far, and we report the mean and standard deviation of two metrics: Final Accuracy (\textsc{FA}), measured at the end of the training stream on all classes, and Continual Accuracy (\textsc{CA}), computed by averaging accuracies evaluated at the end of each task on all classes seen up to that point, reflecting representation quality throughout the stream. Forgetting metrics for all methods are reported in appendix~\ref{app:forgetting}. Statistical significance is assessed by a Student's $t$-test at threshold $p < 0.05$. All baseline results are obtained using the publicly available implementation of~\citet{cignoniCLALatentAlignment2025}, with hyperparameters set as in their original paper.

\subsection{Ablation Study} 
An ablation study is conducted to analyze the respective contribution of each CLIMB component. Always preserving an equivalent CBP, four variants are first evaluated: an SSL-only model ($\mathcal{L}_{\text{SSL}}$), corresponding to plain SimCLR without any continual learning mechanism, to which the hierarchical centroid memory is added successively ($\mathcal{L}_{\text{SSL}}$+Memory), then the alignment loss $\mathcal{L}_{\text{align}}$ computed on replayed examples with the model at the previous training step ($\mathcal{L}_{\text{SSL}}$+Memory+$\mathcal{L}_{\text{align}}$), then with the EMA model ($\mathcal{L}_{\text{SSL}}$+Memory+$\mathcal{L}_{\text{align}}$+EMA), constituting complete CLIMB. To isolate the contribution of the centroid-based memory structure, three additional variants replace it with alternative buffer strategies while keeping all other components identical: a FIFO buffer ($\mathcal{L}_{\text{SSL}}$+FIFO+$\mathcal{L}_{\text{align}}$+EMA), which corresponds to CLA-E~\citep{cignoniCLALatentAlignment2025} trained with CLIMB's learning rate and alignment loss weight $\lambda$, a MinRed buffer~\citep{purushwalkamChallengesContinuousSelfSupervised2022} ($\mathcal{L}_{\text{SSL}}$+MinRed+$\mathcal{L}_{\text{align}}$+EMA), and a reservoir buffer~\citep{vitterRandomSamplingReservoir1985} ($\mathcal{L}_{\text{SSL}}$+Reservoir+$\mathcal{L}_{\text{align}}$+EMA).

\subsection{Model Configuration} 

CLIMB's hyperparameters are configured as follows. The EMA reference model is updated with coefficient $\tau_{\text{ema}} = 0.999$ after each mini-batch, following CLA~\citep{cignoniCLALatentAlignment2025}. The total memory budget is fixed at $N = 2500$ images for all methods. 

The remaining hyperparameters were selected via grid search on the corresponding dataset with 20 tasks. The learning rate, alignment loss weight $\lambda$ and STM centroid EMA coefficient $\alpha_{\text{stm}}$ were searched jointly over $\text{lr} \in \{0.01, 0.05, 0.1, 0.3\}$, $\lambda \in \{0.1, 0.5, 1.0, 2.0, 5.0\}$ and $\alpha_{\text{stm}} \in \{0.05, 0.1, 0.3, 0.5\}$ using SGD. For Split ImageNet-100 ,yielding $\text{lr} = 0.1$, $\lambda = 1.0$ and $\alpha_{\text{stm}} = 0.1$. A separate grid search on Split CIFAR-100 yielded $\text{lr} = 0.3$ and $\lambda = 2.0$, the remaining hyperparameters are shared across both datasets. The hierarchical memory parameters ($M$, $L$, $K$) were then searched under the constraint that total allocated capacity remains consistent with $N = 2500$, over $M \in \{10, 20, 30, 50\}$ images per centroid and STM/LTM centroid capacities ($L$, $K$) accordingly, yielding $M = 30$, $L = 100$, $K = 60$. A sensitivity analysis of the memory architecture and update-rule parameters is provided in Appendix~\ref{app:sensitivity}.


\section{Results}
\label{sec:results}

\subsection{Regular Task Distribution}

Results on the 20, 50, and 100 tasks configurations are presented in Table~\ref{tab:regulier}. On CIFAR-100, CLIMB significantly outperforms all methods in CA on 20 tasks, and remains statistically indistinguishable from CLA-R and MinRed on 50 and 100 tasks, no method significantly dominates in FA across all configurations. The absence of clear separation is consistent with observations from the CLA article~\citep{cignoniCLALatentAlignment2025}, where methods show a more pronounced advantage on complex datasets such as ImageNet-100. However, CLIMB outperforms all OCSSL methods on ImageNet-100 across all configurations, in both Final Accuracy (FA) and Continual Accuracy (CA). The gap with CLA-E, the state-of-the-art method, remains around 2 percentage points on 20 tasks but widens as the number of tasks increases, highlighting a limitation of CLA's FIFO buffer. By favoring recent examples, it tends to progressively forget older distributions as task count grows, whereas CLIMB's hierarchical centroid memory maintains more representative coverage of the entire stream. PCMC, reported without its pretraining phase, remains competitive on CIFAR-100 but lags significantly behind CLIMB on ImageNet-100, despite operating under a substantially higher effective CBP due to its offline sleep phases and unbounded memory. Wall-clock time comparisons across all methods are reported in Appendix~\ref{app:wallclock}.

\begin{table}[t]
\caption{Classification performances of the OCSSL methods on 20, 50, and 100 tasks class-incremental settings.}
\label{tab:regulier}
\begin{center}
\small
\begin{tabular}{clcc|cc}
\toprule
& & \multicolumn{2}{c}{\textbf{CIFAR-100}} & \multicolumn{2}{c}{\textbf{ImageNet-100}} \\
& \textbf{Method} & \textbf{CA} & \textbf{FA} & \textbf{CA} & \textbf{FA} \\
\midrule
& \textit{i.i.d.} & --- & 50.45{\small$\pm$2.03} & --- & 53.93{\small$\pm$1.52} \\
\midrule
\multirow{7}{*}{\rotatebox{90}{20 tasks}}
& SCALE       & 27.88{\small$\pm$1.30} & 31.32{\small$\pm$0.40} & 28.70{\small$\pm$1.45} & 33.43{\small$\pm$0.59} \\
& Osiris-R    & 34.13{\small$\pm$1.29} & 37.65{\small$\pm$0.57} & 37.06{\small$\pm$1.80} & 42.72{\small$\pm$2.00} \\
& MinRed      & 39.34{\small$\pm$1.14} & \textbf{43.89}{\small$\pm$1.44} & 35.87{\small$\pm$1.99} & 43.34{\small$\pm$1.71} \\
& CLA-R       & 39.87{\small$\pm$0.88} & \textbf{42.89}{\small$\pm$1.72} & 42.86{\small$\pm$1.48} & 49.89{\small$\pm$1.12} \\
& CLA-E       & 37.59{\small$\pm$1.14} & 40.95{\small$\pm$0.98} & \textbf{45.52}{\small$\pm$1.22} & \textbf{51.03}{\small$\pm$1.61} \\
& \textbf{CLIMB}       & \textbf{41.33}{\small$\pm$0.72}& \textbf{44.09}{\small$\pm$0.30}  & \textbf{47.46}{\small$\pm$1.76} & \textbf{52.92}{\small$\pm$1.14} \\
\cmidrule(l){2-6}
&PCMC &  \textbf{39.26}{\small$\pm$1.13} & \textbf{43.01}{\small$\pm$0.77}  & 40.11{\small$\pm$1.15} & 39.42{\small$\pm$1.73}\\
\midrule
\multirow{7}{*}{\rotatebox{90}{50 tasks}}
& SCALE       & 27.40{\small$\pm$0.86} & 31.23{\small$\pm$0.50} & 29.26{\small$\pm$3.64} & 31.74{\small$\pm$1.86} \\
& Osiris-R    & 31.94{\small$\pm$1.08} & 37.19{\small$\pm$0.74} & 35.64{\small$\pm$1.80} & 39.02{\small$\pm$0.93} \\
& MinRed      & \textbf{38.14}{\small$\pm$1.61} & \textbf{43.62}{\small$\pm$1.48} & 34.83{\small$\pm$2.29} & 42.70{\small$\pm$1.11} \\
& CLA-R       & \textbf{38.68}{\small$\pm$1.26} & \textbf{41.67}{\small$\pm$1.85} & 40.41{\small$\pm$1.64} & 46.06{\small$\pm$0.78} \\
& CLA-E       & 36.60{\small$\pm$1.38} & 40.65{\small$\pm$1.25} & 43.39{\small$\pm$0.96} & 46.78{\small$\pm$1.30} \\
& \textbf{CLIMB}       & \textbf{38.68}{\small$\pm$1.04} & \textbf{43.15}{\small$\pm$0.58} & \textbf{46.22}{\small$\pm$1.27} & \textbf{50.34}{\small$\pm$0.61} \\
\cmidrule(l){2-6}

&PCMC &  \textbf{38.31}{\small$\pm$1.77} & \textbf{43.13}{\small$\pm$4.16}  & 42.65{\small$\pm$1.54} & 44.46{\small$\pm$2.33} \\
\midrule
\multirow{7}{*}{\rotatebox{90}{100 tasks}}
& SCALE       & 27.25{\small$\pm$1.15} & 31.14{\small$\pm$0.73} & 22.87{\small$\pm$3.09} & 28.08{\small$\pm$1.86} \\
& Osiris-R    & 32.91{\small$\pm$1.43} & 35.48{\small$\pm$1.51} & 35.04{\small$\pm$1.71} & 39.44{\small$\pm$1.47} \\
& MinRed      & \textbf{38.30}{\small$\pm$1.42} & \textbf{43.46}{\small$\pm$1.48} & 35.07{\small$\pm$1.90} & 41.74{\small$\pm$0.83} \\
& CLA-R       & \textbf{38.84}{\small$\pm$1.05} & \textbf{42.28}{\small$\pm$2.26} & 39.57{\small$\pm$1.43} & 45.81{\small$\pm$0.88} \\
& CLA-E       & 36.70{\small$\pm$1.31} & 41.23{\small$\pm$1.27} & 42.20{\small$\pm$1.29} & 46.78{\small$\pm$1.30} \\
& \textbf{CLIMB}       &\textbf{38.60}{\small$\pm$1.22} & \textbf{43.37}{\small$\pm$0.84} & \textbf{44.58}{\small$\pm$1.13} & \textbf{50.21}{\small$\pm$0.40} \\
\cmidrule(l){2-6}
&PCMC & 36.72±1.64 & 39.48±2.79 & 40.53{\small$\pm$1.13} & 38.12{\small$\pm$5.64} \\
\bottomrule
\end{tabular}
\end{center}
\end{table}

\subsection{Irregular Task Distribution}

Results on the irregular task distribution configuration are presented in Table~\ref{tab:irregular}. These experiments are conducted on both Split CIFAR-100 and Split ImageNet-100 and exclude SCALE due to its inferior performance in the regular task distribution experiments. On CIFAR-100, CLIMB, CLA-R, and MinRed are statistically indistinguishable in CA across both configurations, while CLIMB and MinRed lead in FA, with CLA-R significantly below in both configurations. This result is consistent with the regular task distribution setting, confirming that CLIMB achieves competitive performance on CIFAR-100. On ImageNet-100, CLIMB outperforms all OCSSL methods in both evaluated configurations (20 and 50 tasks), in both Final Accuracy and Continual Accuracy. The gap with CLA-E, the most competitive method, is comparable to that observed in the regular configuration, confirming that CLIMB's advantage on ImageNet-100 does not depend on the regularity of the task distribution. MinRed and Osiris-R exhibit markedly inferior performance, suggesting that their less structured replay strategies struggle to cover the diversity of encountered distributions when task fragmentation is variable. PCMC shows no clear advantage in the irregular setting either, achieving results comparable to the weakest OCSSL methods on ImageNet-100.

\begin{table}[t]
\caption{Classification performances of OCSSL methods with irregular task distribution on Split CIFAR-100 and Split ImageNet-100.
}
\label{tab:irregular}
\begin{center}
\small
\begin{tabular}{clcc|cc}
\toprule
& & \multicolumn{2}{c}{\textbf{CIFAR-100}} & \multicolumn{2}{c}{\textbf{ImageNet-100}} \\
& \textbf{Method}& \textbf{CA} & \textbf{FA}  & \textbf{CA} & \textbf{FA} \\
\midrule
\multirow{6}{*}{\rotatebox{90}{20 tasks}}
& Osiris-R            & 33.37{\small$\pm$1.41} & 36.94{\small$\pm$1.77}& 37.52{\small$\pm$2.71} & 40.68{\small$\pm$2.61} \\
& MinRed             & \textbf{38.82}{\small$\pm$1.18} & \textbf{43.42}{\small$\pm$1.97}& 36.47{\small$\pm$2.46} & 42.46{\small$\pm$2.22} \\
& CLA-R              & \textbf{39.27}{\small$\pm$1.36} & 42.42{\small$\pm$2.44} & 43.70{\small$\pm$2.41} & 48.74{\small$\pm$2.65} \\
& CLA-E              & 37.04{\small$\pm$1.42} & 41.16{\small$\pm$1.81}& 46.30{\small$\pm$2.33} & 49.68{\small$\pm$2.87} \\
& \textbf{CLIMB}              & \textbf{39.00}{\small$\pm$1.58 } & \textbf{43.59}{\small$\pm$1.54} & \textbf{47.99}{\small$\pm$2.57} & \textbf{52.19}{\small$\pm$2.55} \\
\cmidrule(l){2-6}
& PCMC & 38.50±1.46 & 39.65±4.75 & 38.44±1.49 & 37.07±4.70 \\
\midrule
\multirow{6}{*}{\rotatebox{90}{50 tasks}}
& Osiris-R           & 32.46{\small$\pm$1.48} & 36.41{\small$\pm$1.38} & 37.35{\small$\pm$1.79} & 40.07{\small$\pm$2.07} \\
& MinRed            & \textbf{38.61}{\small$\pm$1.46} & \textbf{43.32}{\small$\pm$1.20} & 36.99{\small$\pm$1.94} & 42.18{\small$\pm$2.64} \\
& CLA-R              & \textbf{38.89}{\small$\pm$1.62} & 42.40{\small$\pm$1.53}& 42.68{\small$\pm$1.70} & 46.53{\small$\pm$1.70} \\
& CLA-E              & 36.79{\small$\pm$1.65} & 41.03{\small$\pm$1.62} & 45.28{\small$\pm$1.83} & 47.49{\small$\pm$1.70} \\
& \textbf{CLIMB}              & \textbf{38.93}{\small$\pm$1.74} & \textbf{43.22}{\small$\pm$1.68}& \textbf{47.94}{\small$\pm$1.94} & \textbf{50.82}{\small$\pm$1.65} \\
\cmidrule(l){2-6}
& PCMC & 37.35±1.50 & 36.98±2.65 & 40.58±2.31 & 38.85±5.80 \\
\bottomrule
\end{tabular}
\end{center}
\end{table}

\subsection{Ablation Study}

The ablation study results, presented in Table~\ref{tab:ablation}, clearly confirm the contribution of each CLIMB component and highlight the central role of the proposed intelligent memory. First, the SSL-only model achieves limited performance ($19.64\pm1.06$ and $22.63\pm1.22$), confirming that online self-supervised learning, without an explicit information preservation mechanism, suffers strongly from catastrophic forgetting.

Using the hierarchical centroid memory yields a highly significant performance improvement ($46.09\pm1.69$ and $50.84\pm0.95$), demonstrating that the proposed memory alone already effectively stabilizes learned representations in a continual setting. This substantial improvement validates the idea that CLIMB relies first and foremost on a structured memory capable of preserving relevant information over time.

When the alignment loss $\mathcal{L}_{\text{align}}$ is introduced, performance improves further ($46.56\pm1.58$ and $51.45\pm2.22$), confirming that knowledge distillation on replayed examples helps reduce representation drift beyond what memory alone achieves, though this gain is not statistically significant. The choice of reference model, whether the previous training step model or an EMA-updated encoder as described in Section~\ref{sec:training}, has no statistically significant impact ($47.46\pm1.76$ and $52.92\pm1.14$), suggesting that any stable reference suffices in practice. The EMA variant nonetheless achieves the highest absolute performance of all configurations, and we therefore retain it as the reference in the final CLIMB configuration, as it additionally provides a smoother reference trajectory without adding computational overhead.

Finally, replacing the centroid-based memory with alternative buffer strategies while keeping all other components identical yields a significant performance drop in FA, confirming that CLIMB's advantage is attributable to the memory design. Notably, the $\mathcal{L}_{\text{SSL}}$ + Memory variant alone achieves performance statistically equivalent to FIFO, MinRed, and Reservoir buffers combined with distillation, highlighting the importance of memory quality in the OCSSL setting. 

\begin{table}[t]
\caption{Ablation study on Split ImageNet-100, 20 tasks.}
\label{tab:ablation}
\begin{center}
\small
\begin{tabular}{lcc}
\toprule
\textbf{Method} & \textbf{CA} & \textbf{FA} \\
\midrule
$\mathcal{L}_{\text{SSL}}$
    & 19.64{\small$\pm$1.06} & 22.63{\small$\pm$1.22} \\
$\mathcal{L}_{\text{SSL}}$ + Memory
    & \textbf{46.09}{\small$\pm$1.69} & 50.84{\small$\pm$0.95} \\
$\mathcal{L}_{\text{SSL}}$ + Memory + $\mathcal{L}_{\text{align}}$
    & \textbf{46.56}{\small$\pm$1.58} & \textbf{51.45}{\small$\pm$2.22} \\
$\mathcal{L}_{\text{SSL}}$ + Memory + $\mathcal{L}_{\text{align}}$ + EMA \textbf{(CLIMB)}
    &\textbf{47.46}{\small$\pm$1.76} & \textbf{52.92}{\small$\pm$1.14} \\
\hline
        $\mathcal{L}_{\text{SSL}}$ + FIFO + $\mathcal{L}_{\text{align}}$ + EMA
    & \textbf{45.15}{\small$\pm$1.53} & 50.98{\small$\pm$0.96} \\
    $\mathcal{L}_{\text{SSL}}$ + Minred + $\mathcal{L}_{\text{align}}$ + EMA
    & \textbf{45.73}{\small$\pm$1.39} & 50.98{\small$\pm$0.94} \\
    $\mathcal{L}_{\text{SSL}}$ + Reservoir + $\mathcal{L}_{\text{align}}$ + EMA
    & 44.78{\small$\pm$1.82} & 48.64{\small$\pm$2.00} \\

\bottomrule
\end{tabular}
\end{center}
\end{table}


\section{Conclusion and perspectives}
\label{sec:conclusion}

In this paper, we have presented CLIMB, an online continual self-supervised learning method that learns representations from a continuous stream of unlabeled data, without knowledge of task boundaries and under memory constraints. The central contribution of
CLIMB is a hierarchical centroid-based memory, bounded in total number of stored images, designed to maintain a representative set of the latent space of the stream. Combined with replay-based knowledge distillation, this memory enables CLIMB to tackle catastrophic forgetting while maintaining representative coverage of the entire stream.

Experiments on Split ImageNet-100 in 20, 50, and 100 tasks configurations show that CLIMB outperforms state-of-the-art methods. Experiments with irregular task distributions further confirm the robustness of CLIMB across varied learning regimes. Ablation results show that the hierarchical centroid memory alone already matches the performance of flat buffer strategies combined with distillation, and that combining it with knowledge distillation yields a further significant gain in FA, confirming that both components are necessary to achieve the best performance.

Several directions for improvement can be envisioned. The balance between STM and LTM capacity, as well as the centroid merging strategy, could be refined using more informative indicators than cosine similarity alone, incorporating for instance usage frequency. The use of SSL methods that promote a better-structured latent space could reinforce the relevance of centroid groupings and thus improve replay quality. Finally, rather than storing raw images, the memory could preserve compressed representations or leverage a generative model, further reducing memory footprint, although this would need to find a way to tackle the representation drift issue that storing raw images currently avoids.

\subsubsection*{Acknowledgments}
This work was granted access to the HPC resources of IDRIS under the allocation 2025-AD011016434 and 2025-AD011014045R2 granted by GENCI on the V100 partition of the Jean Zay supercomputer. This work was funded by Lyon 1 Université and the Soutien aux ENSeignants-chercheurs (SENS) call for projects.

\bibliography{collas2026_conference}
\bibliographystyle{collas2026_conference}

\appendix

\section{Irregular Task Distribution}
\label{app:distributions}

\begin{figure}[h]
    \centering
    \begin{minipage}{0.48\linewidth}
        \centering
        \includegraphics[width=\linewidth]{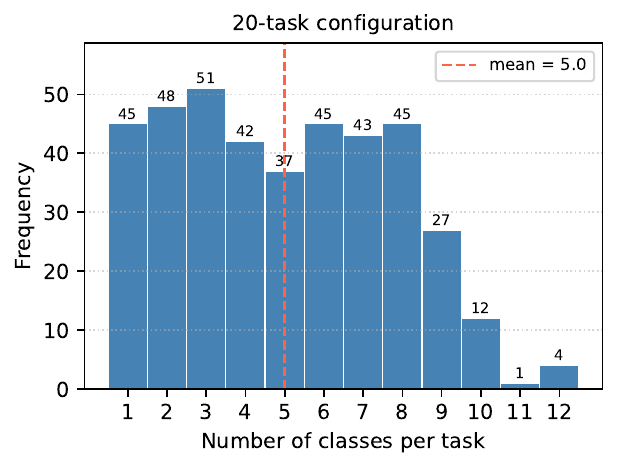}
    \end{minipage}
    \hfill
    \begin{minipage}{0.48\linewidth}
        \centering
        \includegraphics[width=\linewidth]{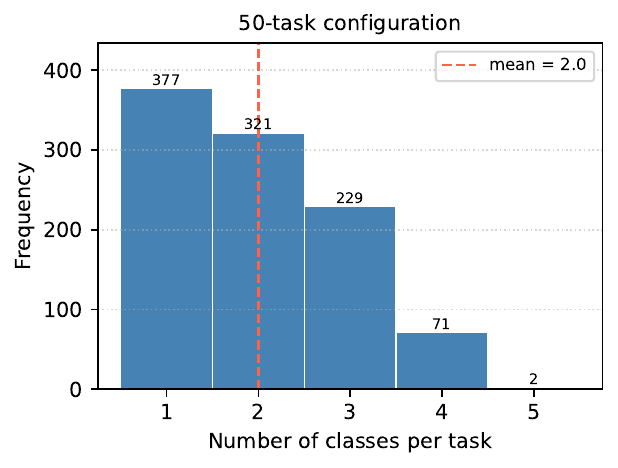}
    \end{minipage}
    \caption{Frequency histogram of the number of classes per task pooled over all 20 seeds, for the 20-task (left) and 50-task (right) configurations. The dashed line indicates the mean value, which corresponds to the fixed number of classes per task in the regular protocol.}
    \label{fig:dist_irreg}
\end{figure}

Figure~\ref{fig:dist_irreg} shows the frequency distribution of the number of classes per task, pooled across all 20 seeds, for the 20-task and 50-task configurations. These distributions are identical for Split CIFAR-100 and Split ImageNet-100, as the same seeds control the class-to-task assignment in both cases.

\section{Hyperparameter Sensitivity Analysis}
\label{app:sensitivity}

\begin{table}[h]
\caption{Sensitivity to memory architecture parameters on Split 
ImageNet-100 (20 tasks, 5 seeds). \underline{Underlined}: reference 
configuration.}
\label{tab:sens_arch}
\begin{center}
\small
\begin{tabular}{llcc}
\toprule
Parameter & Value & \textbf{CA} & \textbf{FA} \\
\midrule
\multirow{6}{*}{$N$ (total images)}
& 1000             & 43.23{\small$\pm$2.61} & 47.39{\small$\pm$1.09} \\
& 1500             & 44.69{\small$\pm$1.84} & 48.55{\small$\pm$0.87} \\
& 2000             & \textbf{45.71}{\small$\pm$2.46} & 50.41{\small$\pm$1.41} \\
& \underline{2500} & \textbf{47.46}{\small$\pm$1.76} & \textbf{52.92}{\small$\pm$1.14} \\
& 3500             & \textbf{47.18}{\small$\pm$1.94} & \textbf{52.78}{\small$\pm$1.02} \\
& 5000             & \textbf{47.05}{\small$\pm$1.98} & \textbf{53.18}{\small$\pm$1.35} \\
\midrule
\multirow{5}{*}{$L$ (STM centroids)}
& 50               & \textbf{45.88}{\small$\pm$1.55} & \textbf{52.03}{\small$\pm$0.78} \\
& 75               & \textbf{46.58}{\small$\pm$1.71} & \textbf{51.70}{\small$\pm$1.10} \\
& \underline{100}  & \textbf{47.46}{\small$\pm$1.76} & \textbf{52.92}{\small$\pm$1.14} \\
& 150              & \textbf{46.13}{\small$\pm$1.65} & 51.12{\small$\pm$1.04} \\
& 200              & \textbf{46.52}{\small$\pm$1.85} & \textbf{52.26}{\small$\pm$1.01} \\
\midrule
\multirow{5}{*}{$K$ (LTM centroids)}
& 20               & 44.97{\small$\pm$1.49} & 50.29{\small$\pm$1.35} \\
& 35               & \textbf{46.31}{\small$\pm$1.87} & 51.58{\small$\pm$0.63} \\
& 50               & \textbf{46.98}{\small$\pm$1.38} & \textbf{52.42}{\small$\pm$0.35} \\
& \underline{60}   & \textbf{47.46}{\small$\pm$1.76} & \textbf{52.92}{\small$\pm$1.14} \\
& 80               & \textbf{45.95}{\small$\pm$1.82} & 50.91{\small$\pm$1.11} \\
\midrule
\multirow{4}{*}{$M$ (images/centroid)}
& 10               & \textbf{45.86}{\small$\pm$1.23} & 51.66{\small$\pm$0.60} \\
& 20               & \textbf{46.00}{\small$\pm$1.89} & 51.09{\small$\pm$1.16} \\
& \underline{30}   & \textbf{47.46}{\small$\pm$1.76} & \textbf{52.92}{\small$\pm$1.14} \\
& 50               & \textbf{46.19}{\small$\pm$1.99} & \textbf{51.32}{\small$\pm$2.03} \\
\bottomrule
\end{tabular}
\end{center}
\end{table}

\begin{table}[h]
\caption{Sensitivity to update-rule parameters on Split ImageNet-100 
(20 tasks, 5 seeds). \underline{Underlined}: reference configuration.}
\label{tab:sens_update}
\begin{center}
\small
\begin{tabular}{llcc}
\toprule
Parameter & Value & \textbf{CA} & \textbf{FA} \\
\midrule
\multirow{5}{*}{Novelty percentile $p$}
& 0.75              & 42.67{\small$\pm$1.33} & 46.88{\small$\pm$0.42} \\
& 0.85              & 43.93{\small$\pm$1.86} & 49.47{\small$\pm$0.42} \\
& 0.90              & \textbf{45.95}{\small$\pm$2.59} & \textbf{51.68}{\small$\pm$1.76} \\
& \underline{0.95}  & \textbf{47.46}{\small$\pm$1.76} & \textbf{52.92}{\small$\pm$1.14} \\
& 0.99              & \textbf{45.98}{\small$\pm$2.09} & 51.24{\small$\pm$1.23} \\
\midrule
\multirow{4}{*}{EMA coefficient $\tau_{\text{ema}}$}
& 0.990              & \textbf{46.39}{\small$\pm$2.24} & 51.38{\small$\pm$1.29} \\
& 0.995              & \textbf{46.44}{\small$\pm$2.13} & 51.26{\small$\pm$1.18} \\
& \underline{0.999}  & \textbf{47.46}{\small$\pm$1.76} & \textbf{52.92}{\small$\pm$1.14} \\
& 0.9995             & \textbf{45.77}{\small$\pm$1.76} & 51.04{\small$\pm$1.56} \\
\midrule
\multirow{5}{*}{STM EMA coeff.\ $\alpha_{\text{stm}}$}
& 0.01              & \textbf{45.40}{\small$\pm$2.41} & \textbf{51.08}{\small$\pm$1.87} \\
& 0.05              & \textbf{45.72}{\small$\pm$2.01} & 50.07{\small$\pm$1.19} \\
& \underline{0.1}   & \textbf{47.46}{\small$\pm$1.76} & \textbf{52.92}{\small$\pm$1.14} \\
& 0.3               & \textbf{45.87}{\small$\pm$1.89} & 51.54{\small$\pm$0.84} \\
& 0.5               & \textbf{46.07}{\small$\pm$1.68} & \textbf{51.78}{\small$\pm$0.88} \\
\bottomrule
\end{tabular}
\end{center}
\end{table}

Tables~\ref{tab:sens_arch} and~\ref{tab:sens_update} report the sensitivity of CLIMB to its memory architecture and update-rule parameters, respectively, on Split ImageNet-100 with 20 tasks over 5 seeds.
\paragraph{Coupling constraints.}
To perform these experiments, we vary the values of $N$, $L$, $K$, $M$, and the update-rule parameters independently. However, some of these memory parameters are not independent under a fixed total budget $N$. When varying $N$, the number of LTM centroids $K$ is adjusted proportionally as $K = \lfloor 60 \times N / 2500 \rceil$ to ensure that the LTM capacity $K \times M$ remains consistent with the total budget, $M$ is kept fixed at 30. When varying $M$, $K$ is adjusted so that the total LTM image capacity $K \times M$ remains approximately constant at 1800, corresponding to the LTM capacity at the reference budget $N = 2500$. The $L$ and $K$ experiments do not require coupling adjustments.
\paragraph{Memory architecture parameters (Table~\ref{tab:sens_arch}).}
Performance increases monotonically with $N$ up to the reference value of 2500, beyond which it levels off, suggesting that $N = 2500$ is sufficient for the memory to maintain adequate stream coverage, noting that the remaining hyperparameters were also tuned at this budget. Regarding $L$, performance is relatively stable across the tested range, with the reference value $L = 100$ achieving the best result, both smaller and larger values lead to slightly lower performance. For $K$, performance peaks at the reference $K = 60$, with $K = 50$ achieving statistically indistinguishable results, suggesting robustness to moderate reductions in LTM capacity. Performance degrades at $K = 80$ in FA, note that due to the dependence between $K$, $M$, and $N$, with $K = 80$ and $M = 30$, only 100 images remain available for the STM out of the total budget $N = 2500$, drastically reducing its staging capacity. For $M$, performance peaks at the reference $M = 30$ and slightly decreases for both smaller and larger values, while remaining statistically comparable in CA, reflecting a trade-off between centroid granularity and per-centroid example diversity.

\paragraph{Update-rule parameters (Table~\ref{tab:sens_update}).} The novelty percentile $p$ controls how selective the threshold is: lower values make the threshold more permissive, creating many small centroids, while higher values are more conservative. As a slight tendency, performance degrades below $p = 0.90$, while $p = 0.90$, $p = 0.95$, and $p = 0.99$ remain statistically comparable, confirming the stability of the method across a wide range of threshold selectivity. Performances are globally similar across all tested update-rule parameters, always statistically indistinguishable in CA and most often in FA, illustrating the robustness of CLIMB to these hyperparameters. As a slight tendency, the EMA coefficient $\tau_{\text{ema}}$ shows that slower update speeds (higher $\tau_{\text{ema}}$) benefit performance up to the reference value of 0.999, and the STM EMA coefficient $\alpha_{\text{stm}}$ exhibits a similar pattern, with the reference value $\alpha_{\text{stm}} = 0.1$ providing the best trade-off between centroid responsiveness and stability.

\section{STM/LTM Memory Dynamics}
\label{app:dynamics}

\begin{figure}[h]
    \centering
    \begin{minipage}{0.48\linewidth}
        \centering
        \includegraphics[width=\linewidth]{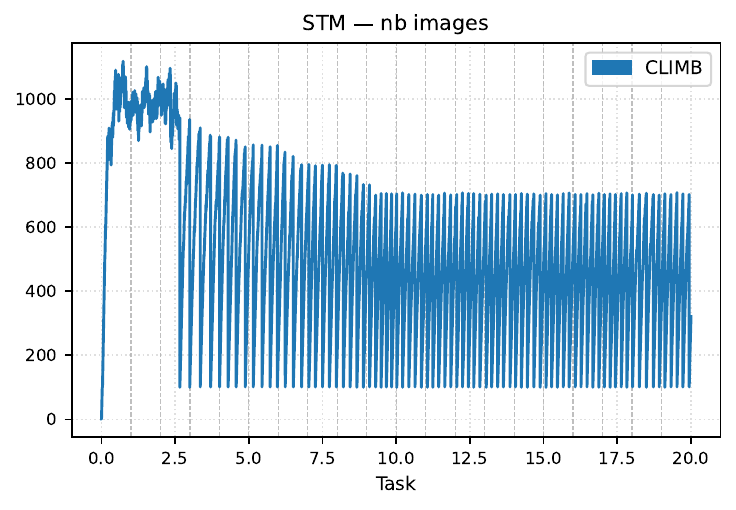}
    \end{minipage}
    \hfill
    \begin{minipage}{0.48\linewidth}
        \centering
        \includegraphics[width=\linewidth]{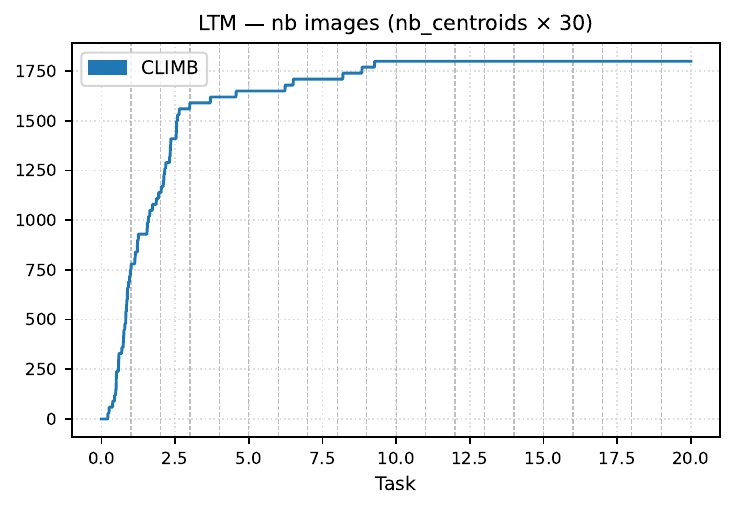}
    \end{minipage}
    \caption{Number of images stored in the STM (left) and LTM (right) as a function of task.}
    \label{fig:dyn_counts}
\end{figure}

\begin{figure}[h]
    \centering
    \begin{minipage}{0.48\linewidth}
        \centering
        \includegraphics[width=\linewidth]{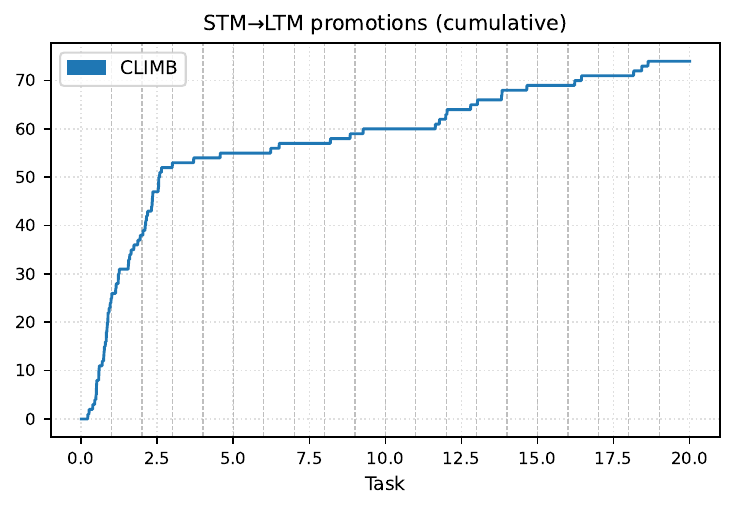}
    \end{minipage}
    \hfill
    \begin{minipage}{0.48\linewidth}
        \centering
        \includegraphics[width=\linewidth]{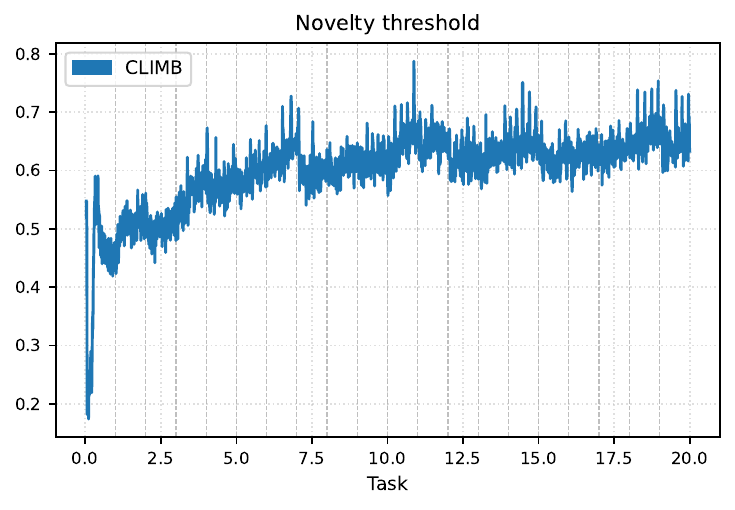}
    \end{minipage}
    \caption{Left: cumulative number of STM$\to$LTM promotions (74 total). Right: adaptive novelty threshold $\tau$, computed as the 95th percentile of the last $w=1000$ observed minimum distances.}
    \label{fig:dyn_promo}
\end{figure}

Figures~\ref{fig:dyn_counts} and ~\ref{fig:dyn_promo} illustrate the internal dynamics of CLIMB's hierarchical memory over a representative run on Split ImageNet-100 with 20 tasks (SimCLR, ResNet-18, seed 1486). Vertical dashed lines mark task boundaries.

The number of images stored in the STM (Figure~\ref{fig:dyn_counts}, left) stabilizes around 700 from task 9 onwards, oscillating between $\sim100$ (immediately after a global pruning event) and $\sim700$ (just before the next pruning). These fluctuations reflect global pruning events triggered when total stored images exceed $N=2500$, which reset STM centroids to a single anchor image each, followed by rapid refilling until the next pruning event. The count of the images in LTM (Figure~\ref{fig:dyn_counts}, right) increases monotonically from 0 and saturates at $K \times M = 60 \times 30 = 1800$ images around task 9, after which every new promotion triggers a merge that maintains the total LTM count at its maximum.

A total of 74 STM$\to$LTM promotions occur over the full run (Figure~\ref{fig:dyn_promo}, left), each corresponding to an STM centroid that accumulated $M=30$ examples. Most promotions occur early in training, before the global pruning mechanism activates between tasks 2 and 3, once pruning begins, STM centroids are regularly reset to a single anchor image, delaying their accumulation toward the promotion threshold and resulting in visible flat segments in the cumulative curve. 

The adaptive novelty threshold $\tau$ (Figure~\ref{fig:dyn_promo}, right) rises during the first seven to ten tasks before stabilizing, oscillating within a narrow range around $0.6$--$0.7$ over the remainder of training. This stabilization reflects that the encoder has built a sufficiently structured latent space: once representations are well-organized, the distribution of minimum distances to existing centroids becomes stationary, and the threshold tracks this stable geometry rather than continuously drifting upward. The early rise corresponds to the period during which the encoder is still learning discriminative features and the representation space is reorganizing.

\section{Global Pruning Strategy Ablation} 
\label{app:trim} 

\begin{table}[h]
\caption{Global pruning strategy ablation on Split ImageNet-100 (5 seeds). \textbf{Anchor}: all STM examples are deleted except one anchor per centroid. \textbf{Selective}: examples are removed from the least recently updated STM centroid until the budget is met.}
\label{tab:trim}
\begin{center}
\small
\begin{tabular}{lccc}
\toprule
\textbf{Strategy} & \textbf{Tasks} & \textbf{CA} & \textbf{FA} \\
\midrule
Anchor   & 20 & \textbf{47.46{\small$\pm$1.76}} & \textbf{52.92{\small$\pm$1.14}} \\
Selective & 20 & \textbf{47.36{\small$\pm$1.35}} & \textbf{52.21{\small$\pm$1.15}} \\
\midrule
Anchor   & 50 & \textbf{46.22{\small$\pm$1.27}} & \textbf{50.34{\small$\pm$0.61}} \\
Selective & 50 & \textbf{45.50{\small$\pm$1.08}} & \textbf{50.19{\small$\pm$1.33}} \\
\bottomrule
\end{tabular}
\end{center}
\end{table}

Table~\ref{tab:trim} compares two strategies for the global pruning step triggered when the total number of stored images exceeds $N$. The \textit{Anchor} strategy, used by default in CLIMB, deletes all STM examples except one anchor image per centroid, retained as a landmark to accommodate future similar samples. The \textit{Selective} strategy instead removes examples incrementally from the least recently updated STM centroid until the budget is met, preserving more recent content at the cost of potentially depleting individual centroids entirely.

Neither strategy is significantly better than the other across both task configurations ($p > 0.05$, Student's $t$-test), confirming that CLIMB is robust to this design choice. The Anchor strategy nonetheless achieves consistently higher absolute performance and is therefore retained as the default.

\section{Memory Diversity Analysis}
\label{app:diversity}

Table~\ref{tab:diversity} compares the memory structure of CLIMB against MinRed, Reservoir, and FIFO at the end of a representative training run on Split ImageNet-100 with 20 tasks. To isolate the contribution of the memory module, all four buffers are evaluated within CLIMB's learning pipeline under identical CBP budget, thus differing only in how images are selected and organized in memory. We report the Vendi score~\citep{friedman2023vendi} as a measure of global memory diversity, the uniformity metric of~\citet{wang2020understanding} as a measure of coverage of the unit hypersphere, and an Inter/Intra ratio measuring the semantic separability of stored examples.

\begin{table}[h]
\caption{Memory diversity metrics at end of training on Split 
ImageNet-100 (20 tasks). All buffers operate under capacity $N=2500$. }
\label{tab:diversity}
\begin{center}
\small
\begin{tabular}{lcccc}
\toprule
\textbf{Metric} & \textbf{CLIMB} & \textbf{MinRed} & \textbf{Reservoir} & \textbf{FIFO} \\
\midrule
Stored images (end of training)   & 2121  & 2500  & 2500  & 2500  \\
\midrule
Vendi score~\citep{friedman2023vendi} $(\uparrow)$
                                   & 43.1  & 35.3  & \textbf{53.3}   & 29.2  \\
Uniformity~\citep{wang2020understanding} $(\downarrow)$
                                   & -3.550 & -3.582 & \textbf{-3.624} & -3.419 \\
\midrule
Intra-cluster dist $(\downarrow)$  & \textbf{0.264} & 0.453{\small$\pm$0.003} & 0.324{\small$\pm$0.002} & 0.288{\small$\pm$0.002} \\
Inter/Intra ratio $(\uparrow)$     & \textbf{3.796} & 2.212{\small$\pm$0.017} & 3.101{\small$\pm$0.020} & 3.485{\small$\pm$0.036} \\
\midrule
Linear probe accuracy $(\uparrow)$ & \textbf{0.510} & 0.491 & 0.485 & 0.501 \\
\bottomrule
\end{tabular}
\end{center}
\end{table}

The Vendi score is defined as $\mathrm{VS} = \exp(-\operatorname{tr}(\mathbf{K}/n \log \mathbf{K}/n))$, where $\mathbf{K} \in \mathbb{R}^{n \times n}$ is the cosine similarity matrix computed over the encoder embeddings of all $n$ images stored in the buffer~\citep{friedman2023vendi}. Intuitively, VS can be read as the effective number of distinct elements in the buffer, ranging from $1$ when all embeddings are identical to $n$ when all are mutually orthogonal. The uniformity metric is $\mathcal{L}_{\mathrm{uniform}} = \log \mathbb{E}_{x,y}[e^{-t\|f(x) -f(y)\|^2}]$ with $t=2$, where $f$ denotes the frozen encoder and the expectation is over all pairs of buffered images, more negative values indicate more uniform coverage of the unit hypersphere~\citep{wang2020understanding}. The Inter/Intra ratio  is obtained by running $k$-means with $k=100$, matching the number of classes in the dataset, on a subsample of $2121$ encoder embeddings drawn from each buffer, matching CLIMB's image count at end of training to ensure comparability, since cluster metrics depend on sample size, repeated 10 times for the baselines to account for sampling variance, reported values are means and standard deviations over these repetitions. Higher values indicate clusters that are simultaneously tight and well-separated.

Reservoir achieves the highest raw diversity, both in Vendi score ($53.3$) and uniformity ($-3.624$), yet obtains the lowest linear probe accuracy ($0.485$). This dissociation indicates that raw diversity alone is not the relevant criterion for replay quality: without structural organization, the stored examples do not provide the semantically coherent and well-separated groups that benefit contrastive learning.

FIFO achieves the worst uniformity score ($-3.419$) and the lowest Vendi score ($29.2$), consistent with its buffer being concentrated on the most recent task only. Its competitive Inter/Intra ratio ($3.485$) should reflects locally coherent clusters restricted to recent data.

CLIMB achieves the best Inter/Intra ratio ($3.796$), meaning its clusters are simultaneously internally coherent (intra-cluster distance $0.264$) and mutually well-separated, while maintaining the second-highest Vendi score ($43.1$) despite storing fewer images than MinRed ($35.3$) and FIFO ($29.2$). This combination of local structure and global stream coverage is consistent with the highest linear probe accuracy ($0.510$), suggesting that structured diversity, semantically coherent and well-separated clusters covering the full stream, may be more beneficial for replay quality than raw diversity alone.


\section{Wall-Clock Time}
\label{app:wallclock}

\begin{figure}[h]
    \centering
    \includegraphics[width=0.6\textwidth]{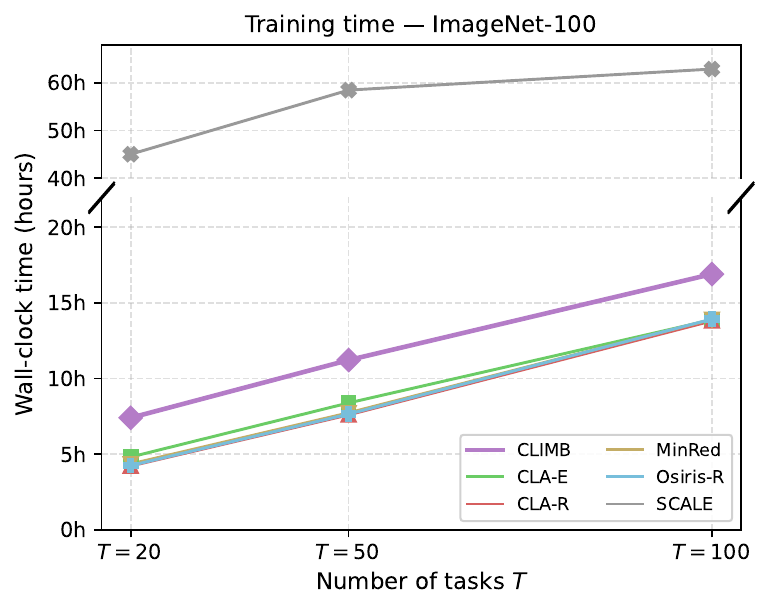}
    \caption{Total training time (hours, log scale) on Split 
    ImageNet-100 as a function of the number of tasks, measured 
    on a single NVIDIA V100 32GB GPU.}
    \label{fig:wallclock}
\end{figure}

Figure~\ref{fig:wallclock} reports the total training time in hours for each method on Split ImageNet-100, measured on a single NVIDIA V100 32GB GPU across the 20, 50, and 100 task configurations, using identical seeds for all methods.

All methods scale quasi-linearly with the number of tasks, a growth driven primarily by the increasing number of linear classifier evaluations: across the 20, 50, and 100 task configurations, the total number of training images is identical, distributed differently across tasks, so the additional cost comes almost entirely from the growing number of evaluation steps performed at task boundaries. CLA-R, Osiris-R, MinRed, and CLA-E form a tight cluster ranging from 4 to 14 hours across all configurations. CLIMB presents a moderate but constant overhead relative to this cluster: the gap with CLA-E remains stable at approximately 2.6 to 3.0 hours regardless of the number of tasks. SCALE constitutes a clear outlier, requiring 3 to 4 times more training time than CLIMB, likely due to its costly memory update procedure.

\section{CLIMB with SimSiam}
\label{app:simsiam}

\begin{table}[h]
\caption{Classification performances with SimSiam on Split CIFAR-100 
and Split ImageNet-100 (5 seeds).}
\label{tab:simsiam}
\begin{center}
\small
\begin{tabular}{clcc|cc}
\toprule
& & \multicolumn{2}{c}{\textbf{CIFAR-100}} & \multicolumn{2}{c}{\textbf{ImageNet-100}} \\
& \textbf{Method} & \textbf{CA} & \textbf{FA} & \textbf{CA} & \textbf{FA} \\
\midrule
& \textit{i.i.d.} & --- & 43.12{\small$\pm$0.56} & --- & 49.75{\small$\pm$0.98} \\
\midrule
\multirow{4}{*}{\rotatebox{90}{20 tasks}}
& MinRed         & \textbf{33.11}{\small$\pm$1.12} & \textbf{37.34}{\small$\pm$2.50} & 34.58{\small$\pm$2.33} & \textbf{42.72}{\small$\pm$1.78} \\
& CLA-R          & \textbf{32.28}{\small$\pm$1.61} & 33.21{\small$\pm$3.79}          & 36.49{\small$\pm$1.56} & 38.99{\small$\pm$4.65} \\
& CLA-E          & 31.01{\small$\pm$0.88}          & 33.31{\small$\pm$3.52}          & 34.23{\small$\pm$2.46} & 41.25{\small$\pm$1.72} \\
& \textbf{CLIMB} & 31.62{\small$\pm$1.34}          & \textbf{35.88}{\small$\pm$1.00} & \textbf{39.21}{\small$\pm$1.75} & \textbf{44.40}{\small$\pm$1.30} \\
\midrule
\multirow{4}{*}{\rotatebox{90}{50 tasks}}
& MinRed         & \textbf{32.57}{\small$\pm$1.42} & \textbf{38.49}{\small$\pm$0.81} & 33.88{\small$\pm$2.32} & \textbf{41.61}{\small$\pm$1.46} \\
& CLA-R          & \textbf{31.80}{\small$\pm$1.55} & 32.45{\small$\pm$4.14}          & \textbf{35.39}{\small$\pm$2.31} & \textbf{39.82}{\small$\pm$1.94} \\
& CLA-E          & 29.58{\small$\pm$1.17}          & 31.57{\small$\pm$3.77}          & 33.48{\small$\pm$2.42} & \textbf{41.11}{\small$\pm$1.21} \\
& \textbf{CLIMB} & \textbf{31.35}{\small$\pm$1.15} & 37.04{\small$\pm$0.78}          & \textbf{37.51}{\small$\pm$2.13} & \textbf{41.79}{\small$\pm$1.97} \\
\bottomrule
\end{tabular}
\end{center}
\end{table}

Tables~\ref{tab:simsiam} report the performance of CLIMB and the main baselines using SimSiam~\citep{chenExploringSimpleSiamese2020} as the base SSL method on Split CIFAR-100 and Split ImageNet-100 respectively. Hyperparameters for CLIMB were selected via grid search on each dataset independently, yielding $\text{lr} = 0.2$ and $\lambda = 0.1$ in both cases. Hyperparameters for all baselines correspond to their best-performing configurations.

On Split CIFAR-100, all methods remain far below the i.i.d.\ upper bound, and CLA-R and CLA-E exhibit high variance in FA (standard deviations of 3--4\%), suggesting instability of SimSiam under these methods' distillation dynamics. CLIMB achieves competitive CA alongside MinRed and CLA-R at both 20 and 50 tasks, and ranks among the top methods in FA at 20 tasks.

On Split ImageNet-100, CLIMB consistently outperforms all other methods in CA at both 20 and 50 tasks, and leads in FA at 20 tasks alongside MinRed, mirroring the pattern observed under SimCLR. At 50 tasks, all methods are statistically indistinguishable in FA while CLIMB and CLA-R form the top group in CA. The consistency between the two SSL methods suggests that CLIMB's advantage stems from its memory structure rather than from properties specific to contrastive learning.

\section{Forgetting Metric}
\label{app:forgetting}

\begin{table}[h]
\caption{Forgetting metric (\%) on Split CIFAR-100 and Split 
ImageNet-100 (5 seeds, lower is better).}
\label{tab:forgetting}
\begin{center}
\small
\begin{tabular}{clcc}
\toprule
& \textbf{Method} & \textbf{CIFAR-100} & \textbf{ImageNet-100} \\
\midrule
\multirow{6}{*}{\rotatebox{90}{20 tasks}}
& SCALE    & $\mathbf{-4.03}${\small$\pm$0.84} & $\mathbf{-4.46}${\small$\pm$1.87} \\
& Osiris-R & $\mathbf{-3.79}${\small$\pm$0.91} & $-4.26${\small$\pm$1.14} \\
& MinRed   & $\mathbf{-4.24}${\small$\pm$1.04} & $\mathbf{-5.64}${\small$\pm$0.65} \\
& CLA-R    & $-1.92${\small$\pm$1.70}           & $-3.06${\small$\pm$0.71} \\
& CLA-E    & $-2.47${\small$\pm$1.18}           & $-1.62${\small$\pm$1.09} \\
& \textbf{CLIMB} & $\mathbf{-3.26}${\small$\pm$0.86} & $-3.06${\small$\pm$1.12} \\
\midrule
\multirow{6}{*}{\rotatebox{90}{50 tasks}}
& SCALE    & $\mathbf{-3.30}${\small$\pm$3.03} & $-2.96${\small$\pm$2.78} \\
& Osiris-R & $\mathbf{-4.59}${\small$\pm$0.88} & $-2.39${\small$\pm$1.02} \\
& MinRed   & $\mathbf{-5.16}${\small$\pm$1.01} & $\mathbf{-7.29}${\small$\pm$1.12} \\
& CLA-R    & $-2.42${\small$\pm$1.70}           & $-4.12${\small$\pm$1.45} \\
& CLA-E    & $\mathbf{-3.52}${\small$\pm$1.74} & $-1.74${\small$\pm$1.61} \\
& \textbf{CLIMB} & $\mathbf{-4.37}${\small$\pm$0.70} & $-3.66${\small$\pm$0.95} \\
\midrule
\multirow{6}{*}{\rotatebox{90}{100 tasks}}
& SCALE    & $\mathbf{-4.58}${\small$\pm$1.32} & $-3.74${\small$\pm$2.96} \\
& Osiris-R & $\mathbf{-4.39}${\small$\pm$1.60} & $-3.62${\small$\pm$1.58} \\
& MinRed   & $\mathbf{-5.57}${\small$\pm$1.87} & $\mathbf{-7.01}${\small$\pm$1.06} \\
& CLA-R    & $-3.25${\small$\pm$1.98}           & $-5.37${\small$\pm$1.59} \\
& CLA-E    & $\mathbf{-4.35}${\small$\pm$1.28} & $-3.14${\small$\pm$1.87} \\
& \textbf{CLIMB} & $\mathbf{-4.66}${\small$\pm$0.60} & $-5.13${\small$\pm$0.58} \\
\bottomrule
\end{tabular}
\end{center}
\end{table}

Table~\ref{tab:forgetting} reports the forgetting metric for all methods, defined as:
\begin{equation}
\text{Forgetting} = \frac{1}{T-1} \sum_{i=1}^{T-1} (a_{i,i} - a_{T,i})
\end{equation}
where $a_{i,j}$ is the evaluation of task $j$ after learning task $i$, and $a_{T,i}$ its accuracy at the end of the stream. Negative values indicate that representations of past tasks improved over training rather than degraded. All methods exhibit negative forgetting across all configurations, indicating that representations of past tasks continue to improve throughout training rather than degrade. This suggests that replay not only prevents catastrophic forgetting but also enables backward transfer, where learning new tasks benefits the representations of previously seen ones. Note that a strongly negative forgetting score does not imply superior overall performance: it reflects a large accuracy gain on past tasks between their initial and final evaluation, but says nothing about the absolute level of those final accuracies. A method that starts from a low initial accuracy on each task and improves substantially over time may exhibit strong negative forgetting while still achieving lower CA and FA than a method that learns better representations from the start.

\end{document}

%% file: math_commands.tex
\usepackage{amsmath,amsfonts,bm}

\def\eqref#1{equation~\ref{#1}}

\def\1{\bm{1}}

\DeclareMathAlphabet{\mathsfit}{\encodingdefault}{\sfdefault}{m}{sl}
\SetMathAlphabet{\mathsfit}{bold}{\encodingdefault}{\sfdefault}{bx}{n}

